%% file: main.tex
\documentclass[]{wechat}

\input{math_commands.tex}

\usepackage{url}
\usepackage{threeparttable}
\usepackage[table]{xcolor} 
\usepackage[bottom]{footmisc} 

\usepackage{CJKutf8}

\usepackage{amsmath,amssymb}
\usepackage{booktabs}
\usepackage{tabularx}
\usepackage{array}
\usepackage{graphicx}
\usepackage{xcolor}
\usepackage{bbm}
\graphicspath{{figure/figs/}}
\usepackage{threeparttable}
\usepackage{enumitem}
\usepackage{float}

\newcolumntype{Y}{>{\centering\arraybackslash}X}

\newcommand{\linkicon}[1]{\raisebox{-0.2em}{\includegraphics[height=1em]{#1}}}
\newcommand{\hficon}{\linkicon{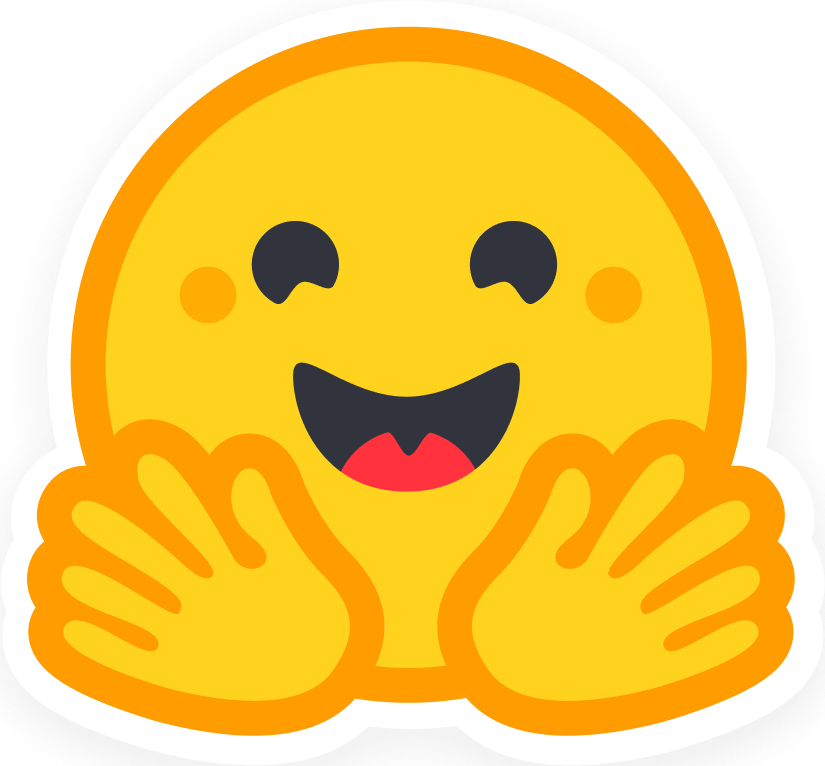}}
\newcommand{\ghicon}{\linkicon{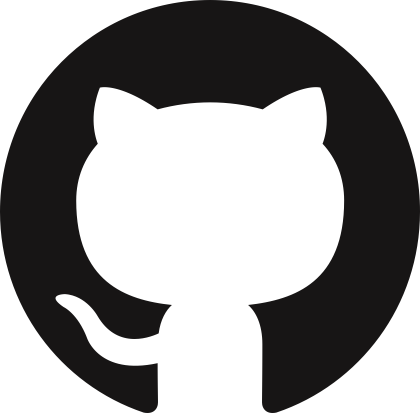}}
\newcommand{\hflink}{https://huggingface.co/collections/tencent/wemm-video}
\newcommand{\ghlink}{https://github.com/Tencent/WeMM-Video}
\newcommand{\resourcelinks}{%
  \begingroup
  \small\raggedright
  \setlength{\parindent}{0pt}%
  \setlength{\parskip}{0pt}%
  \href{\hflink}{\makebox[1.6em][l]{\hficon}\texttt{\hflink}}\par
  \href{\ghlink}{\makebox[1.6em][l]{\ghicon}\texttt{\ghlink}}\par
  \endgroup
}

\title{WeMM-Video: Pushing the Frontiers of Open-Source Omni-Modal Video Understanding} 
\author{%
\fontsize{11.2pt}{13.2pt}\selectfont
\begingroup
\fontsize{10.8pt}{16.5pt}\selectfont
\centerline{%
Changli Tang\textsuperscript{1,2,*},\enspace
Jie Yang\textsuperscript{1,*},\enspace
Tianyi Wang\textsuperscript{1,*,$^{\dagger}$},\enspace
Lei Li\textsuperscript{1,*},\enspace
Daoze Zhang\textsuperscript{1,*},\enspace
Zijie Xin\textsuperscript{1,*},
}
\centerline{%
Zhan Liu\textsuperscript{1,2},\enspace
Li Li\textsuperscript{1},\enspace
Jinchang Xu\textsuperscript{1},\enspace
Fengyun Rao\textsuperscript{1,$^{\ddagger}$},\enspace
Chao Zhang\textsuperscript{2},\enspace
Jing LYU\textsuperscript{1}
}
\endgroup
\vspace{6pt}
\centerline{
   \textsuperscript{1} WeChat Vision, Tencent Inc.
   \textsuperscript{2} Tsinghua University \quad
}
\vspace{-24pt}
}

\vspace{-15pt}
\abstract{

Video understanding, particularly with accompanying audio, remains a central challenge in multimodal intelligence. In this report, we introduce \textbf{WeMM-Video}, a family of Chinese-English bilingual omni-modal foundation models spanning 2B, 4B, 9B, and 35B-A3B variants. Centering on joint audio-visual perception and reasoning, WeMM-Video unifies text, image, audio, and video processing within a hybrid-thinking paradigm, seamlessly supporting both a non-thinking instruct mode for fast, direct responses and a thinking mode for explicit step-by-step reasoning. 
The models are trained on a large-scale multimodal corpus of approximately 1.2 trillion tokens through progressive pre-training and a multi-stage post-training pipeline incorporating multi-teacher OPD, anti-repetition MrDPO, and specialist RL.
Evaluated across more than 120 benchmarks, WeMM-Video demonstrates superior performance in omni-modal video understanding and reasoning, while maintaining highly competitive capabilities across general audio, silent video, image, and text tasks. Specifically, in omni-modal video evaluations, our flagship WeMM-Video-35B-A3B establishes a new open-source state of the art, substantially outperforming proprietary models such as Qwen3.5-Omni-Flash and Seed2.0-Mini-0428 while narrowing the gap with the strong Gemini-3.0-Flash, and our compact WeMM-Video-9B surpasses the larger open-source Qwen3-Omni-30B-A3B-Instruct. We publicly release the WeMM-Video family along with our evaluation and reproduction pipeline to foster further research and development in the open-source community.

\par\vspace{5pt}
\resourcelinks
\vspace{-3.5pt}
}
\date{November 19, 2026}

\begin{document}
\maketitle

\begingroup
\renewcommand{\thefootnote}{}
\footnotetext{\scriptsize
$^{*}$Core contributors.
\quad
$^{\dagger}$Project leader.
\quad
$^{\ddagger}$Corresponding author.
}
\endgroup

\vspace{-22pt}

\begin{figure}[ht]
    \centering
    \includegraphics[width=\linewidth]{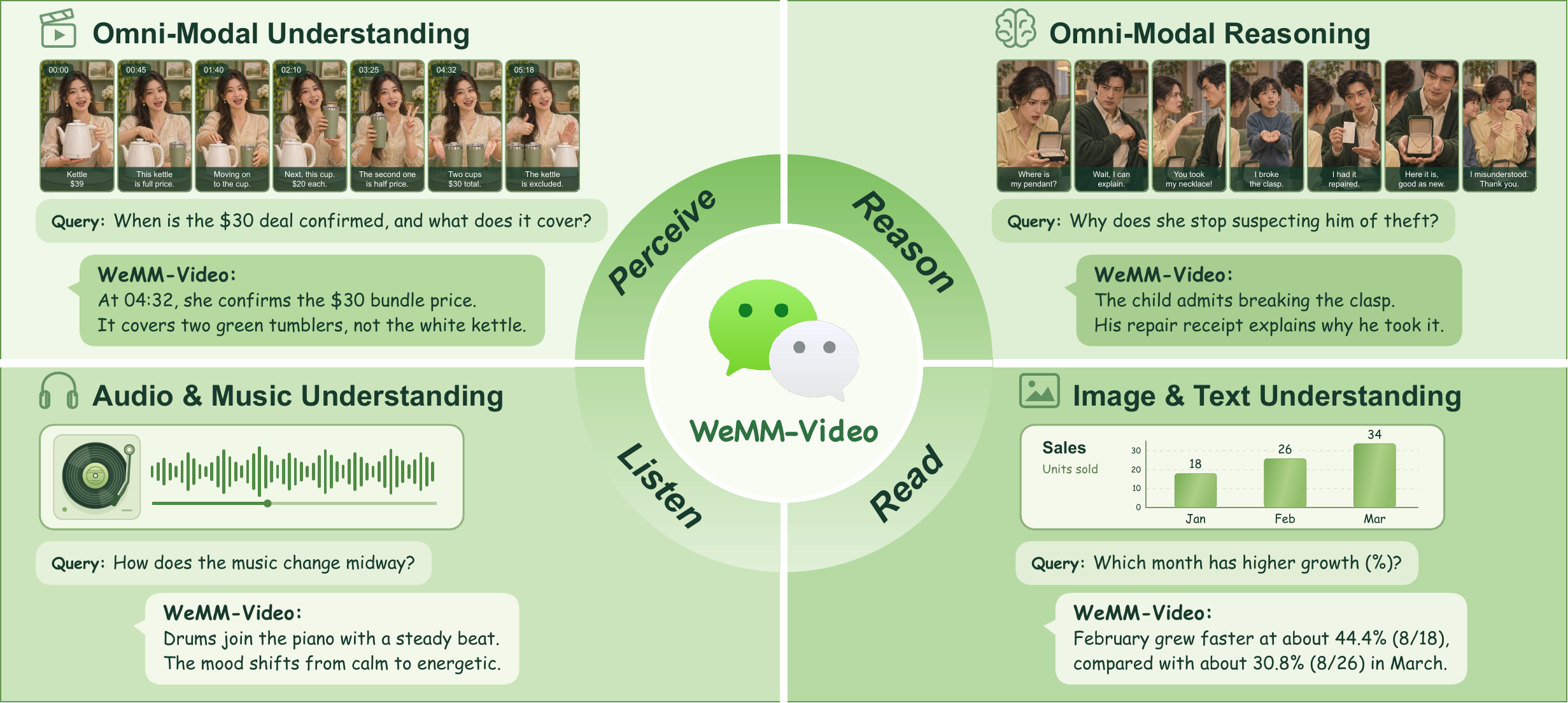}
    \caption{WeMM-Video can perceive and understand omni-modal videos, perform complex multimodal reasoning, and effectively handle general tasks across audio, image, and text modalities.}
    \label{fig:begin}
\end{figure}

\section{Introduction}

Video understanding, particularly the unified comprehension of continuous visual scenes and acoustic signals, represents a pivotal frontier in modern multimodal intelligence. Driven by increasingly prevalent real-world scenarios such as short-form video feeds and embodied intelligence, the demand for machines to perceive dynamic audio-visual environments accurately has grown substantially. Building upon rapid advances in large language models (LLMs) \cite{grattafiori2024llama, singh2025openai, comanici2025gemini, claude3, team2025qwen3, guo2025deepseek, team2026kimi, zeng2026glm}, the multimodal community has witnessed the rise of capable vision-language models \cite{li2023blip, liu2023visual, wang2025internvl3, bai2025qwen3, team2026kwai, an2026llava} and audio LLMs \cite{tang2024salmonn, chu2024qwen2, ding2025kimi, yang2026moss, lin2026stepaudio}, with pioneering efforts beginning to explore collaborative audio-visual capabilities \cite{qwen2.5omni, cui2026minicpm, team2026qwen3, seed2, gemini4argon}.


Despite this progress, comprehensively interpreting real-world videos remains a formidable challenge. Beyond recognizing isolated objects, actions, or sounds, an effective model must capture fine-grained temporal dynamics across frames, connect long-range events, and synthesize visual and acoustic cues through deliberate reasoning to grasp both high-level semantics and subtle details. Existing open-source models, however, remain brittle in complex real-world scenarios: they often overlook critical audio-visual details, fail to align acoustic cues with their corresponding visual events, and suffer from cross-modal hallucinations unsupported by video evidence.

To bridge this gap, we introduce and open-source \textbf{WeMM-Video}, a family of Chinese-English bilingual omni-modal foundation models spanning 2B, 4B, 9B, and 35B-A3B variants. 
Centering on omni-modal video perception and reasoning, WeMM-Video unifies text, image, audio, and video processing within a hybrid-thinking paradigm, supporting both a non-thinking (instruct) mode for low-latency, direct responses and a thinking mode for extended chain-of-thought reasoning. By generating grounded reasoning traces over both visual-only and audio-visual inputs, the thinking mode effectively mitigates cross-modal hallucinations and substantially boosts performance on complex video tasks. To endow the models with these capabilities, we train WeMM-Video on a curated multimodal corpus comprising approximately 1.2 trillion (T) tokens via progressive pre-training, followed by a three-stage post-training pipeline incorporating multi-teacher on-policy distillation (OPD), anti-repetition multi-round direct preference optimization (MrDPO) \cite{tang2025video}, and specialist reinforcement learning (RL).
Across comprehensive omni-modal video benchmarks, our strongest model, WeMM-Video-35B-A3B, achieves state-of-the-art (SOTA) performance among open-source models of comparable scale and rivals leading proprietary models, including Gemini-3.0-Flash, Seed2.0-Mini-0428 and Qwen3.5-Omni-Flash. 
WeMM-Video models also maintain strong, well-rounded capabilities across standalone audio, visual-only video, image, and text tasks. 
Beyond standard benchmarks, WeMM-Video has been deployed across multiple core video services within WeChat, supporting diverse applications including live-stream commerce, short-form drama analysis, and interactive video scenarios.
Recognizing that standardized evaluation and transparent reproducibility are vital for the open-source community, we release our evaluation and reproduction pipeline alongside the model weights. 
We hope the WeMM-Video family can serve as a versatile foundation for community-driven research and real-world applications, pushing the frontiers of open-source omni-modal video understanding.

\begin{figure}[th]
    \centering
    \includegraphics[width=0.95\linewidth]{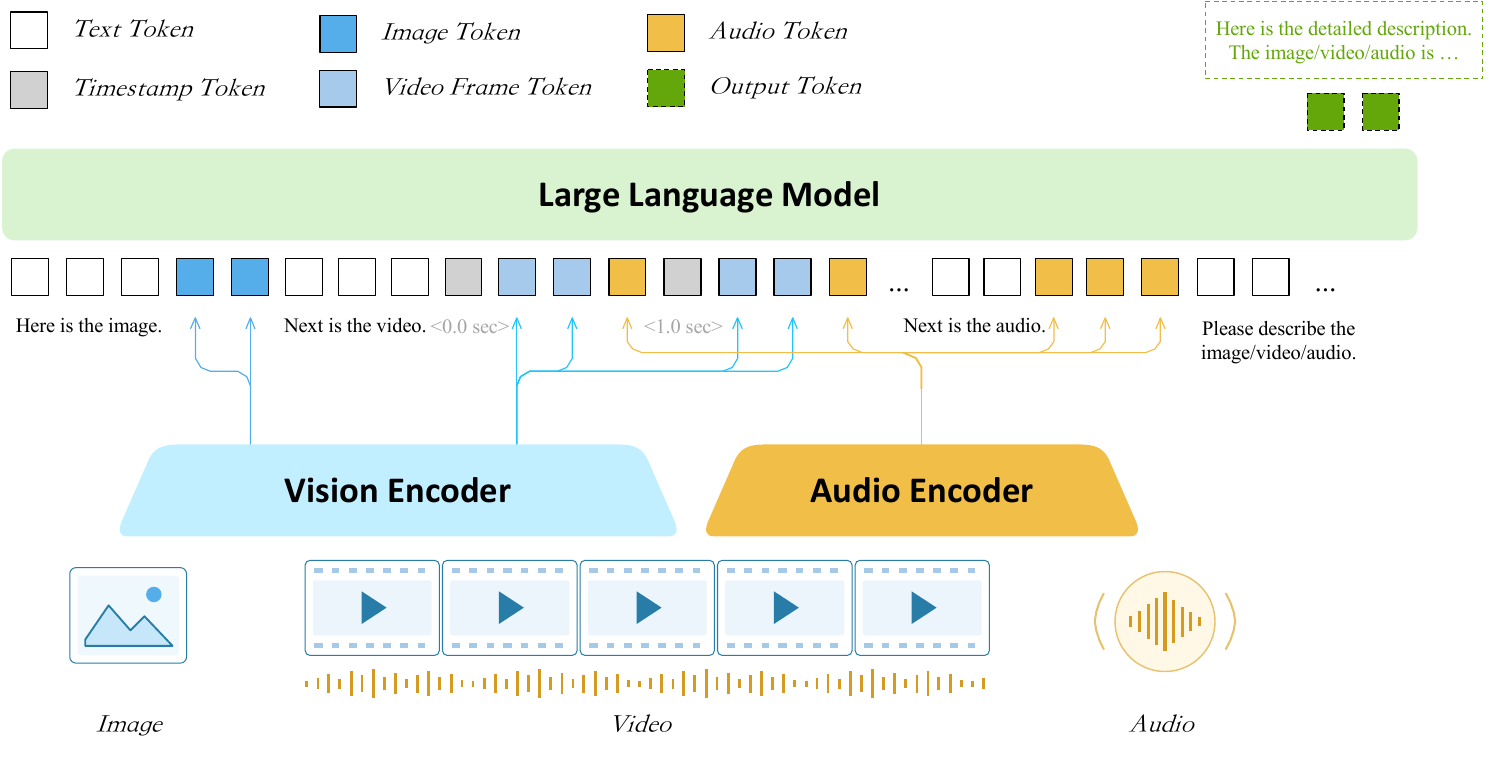}
    \caption{Model architecture of WeMM-Video.}
    \label{fig:arch}
\end{figure}

\section{Model Architecture}
The overall architecture of WeMM-Video primarily draws upon the design of video-SALMONN 2 \cite{tang2025video}, receiving multimodal inputs comprising text, image, audio, and video, and producing textual outputs autoregressively, as illustrated in Fig.~\ref{fig:arch}. 
WeMM-Video-2B, WeMM-Video-4B, and WeMM-Video-9B are built on Qwen3.5-2B, Qwen3.5-4B, and Qwen3.5-9B \cite{qwen3.5_9b}, respectively, while WeMM-Video-35B-A3B is built on Qwen3.6-35B-A3B \cite{qwen36_35b_a3b}. The audio encoders of WeMM-Video models are initialized from Qwen2.5-Omni \cite{qwen2.5omni}.

For visual processing, including both static images and dynamic video frames, we retain the native visual encoders and projection mechanisms of the respective backbones, preserving their pretrained image and video understanding capabilities. 
To handle continuous acoustic signals, WeMM-Video incorporates a dedicated audio module comprising a Whisper-based audio encoder \cite{radford2023robust} and a window-level Q-Former \cite{tang2024extending}. The Whisper-based encoder extracts acoustic features from audio inputs sampled at 16,000~Hz and segmented into 30-second chunks. The window-level Q-Former then compresses these features within local temporal windows of 0.1 seconds across all model variants, mapping each window to a single audio token in the LLM input space and yielding 10 audio tokens per second. For video inputs with accompanying audio, visual and audio tokens are interleaved in chronological order, supplemented with explicit textual timestamps to facilitate fine-grained cross-modal temporal alignment.

\section{Data}
\begin{figure*}[ht]
    \centering
    \begin{subfigure}[b]{0.36\linewidth}
        \centering
        \includegraphics[width=0.667\linewidth]{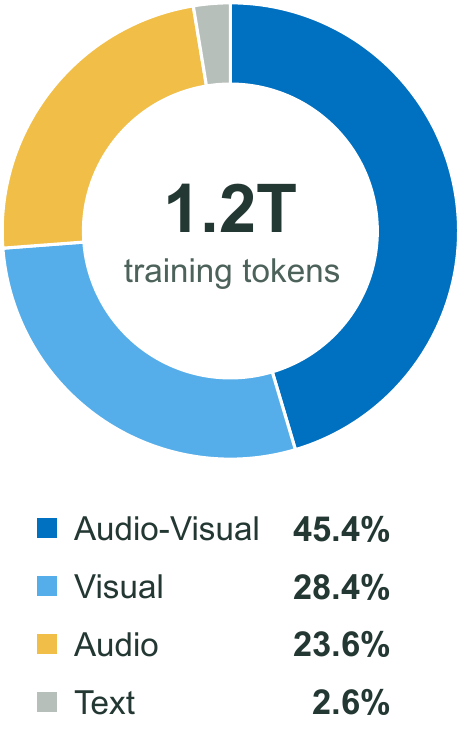}
        \caption{Token distribution by modality.}
        \label{fig:wemm-tokens}
    \end{subfigure}%
    \hspace{0.015\linewidth}%
    \begin{subfigure}[b]{0.47\linewidth}
        \centering
        \includegraphics[width=\linewidth]{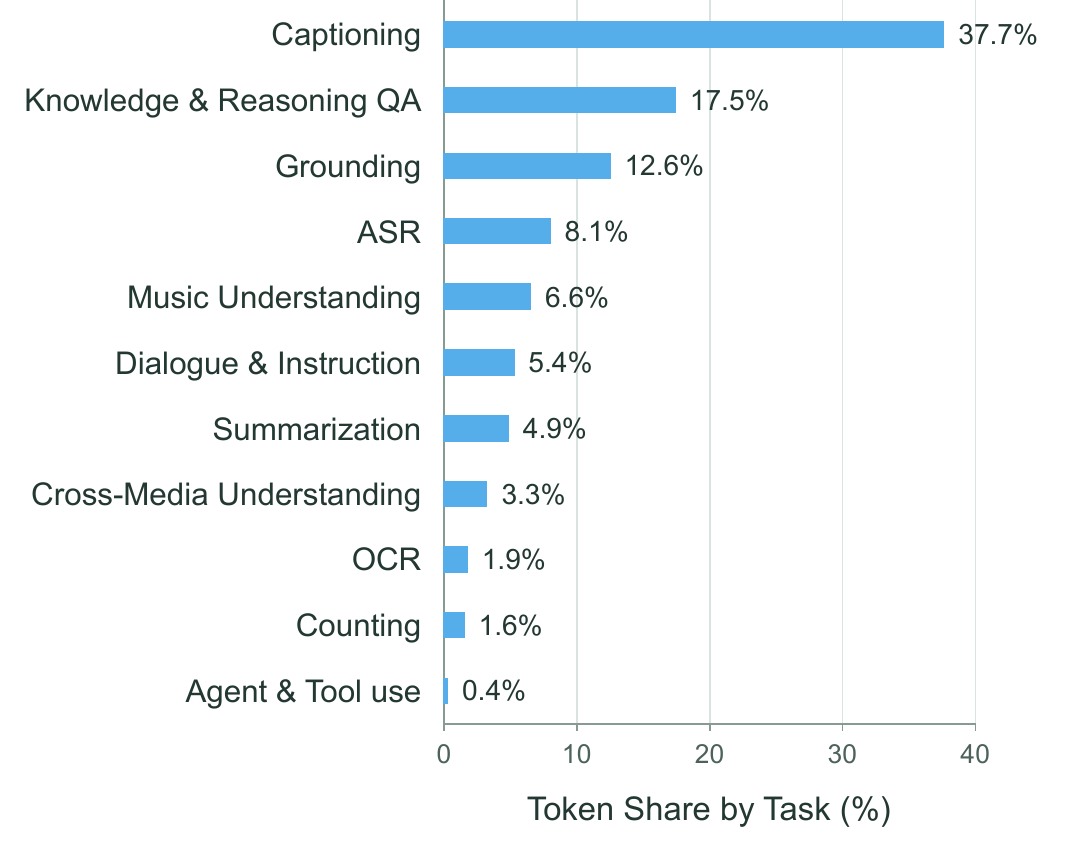}
        \caption{Token distribution by task.}
        \label{fig:wemm-tasks}
    \end{subfigure}
    \caption{Composition of the WeMM-Video training data.}
    \label{fig:wemm-pretraining-data}
\end{figure*}

To establish robust multimodal perception and reasoning, WeMM-Video is trained on a comprehensive, high-capacity training corpus. As shown in Fig.~\ref{fig:wemm-tokens}, the corpus encompasses about 1.2T tokens. Notably, joint audio-visual data constitutes the largest share, contributing 45.4\% of the overall volume, providing dense temporal supervision for modeling complex multimodal dynamics.

Besides, Fig.~\ref{fig:wemm-tasks} illustrates the detailed token distribution across tasks. Captioning accounts for the largest task proportion (37.7\%) and serves as the bedrock for foundational multimodal perception. Within this category, we place strategic emphasis on two complementary types of video captions: fine-grained visual captions and joint audio-visual captions. First, for fine-grained visual captioning, we curate an action- and event-dense video split spanning competitive sports, esports, gaming, choreography, and outdoor activities, annotated with dense descriptions that capture rapid physical dynamics and subtle temporal state transitions. Second, for audio-visual captioning, we scale up native omni-modal videos such as films, short-form dramas, interviews, variety shows, and live-stream commerce, synthesizing comprehensive captions that jointly describe visual actions, spoken dialogue, and ambient acoustic signals along a continuous timeline. All candidate captioning data undergo a rigorous pipeline of automated quality scoring, semantic deduplication, and toxicity filtering to ensure both perceptual fidelity and data diversity.


Since the base model natively lacks acoustic perception, we incorporate a dedicated audio corpus to construct and align the acoustic representation space. This collection covers large-scale automatic speech recognition (ASR), music understanding, and general audio understanding. 

Beyond standard single-video streams, real-world applications frequently require models to process streaming feeds and multi-video contexts. To enhance generalization over heterogeneous and complex inputs, we construct a dedicated cross-media understanding split, targeting multi-source tasks such as pairwise video comparison, video-soundtrack compatibility analysis, and e-commerce product-to-ad consistency verification.


While perceptual understanding is critical, audio-visual joint reasoning remains a significant bottleneck that existing open-source models rarely address effectively. High-quality knowledge, STEM, and complex reasoning data are indispensable for cultivating these advanced cognitive capabilities; however, verifiable audio-visual reasoning data is naturally scarce in the wild. To bridge this gap, we curate a substantial knowledge and reasoning QA collection via a dual strategy: we incorporate large-scale vision-language STEM and knowledge reasoning datasets augmented with high-quality thinking traces, alongside a curated set of native audio-visual reasoning tasks. This synergy effectively transfers deep reasoning capabilities to omni-modal scenarios while preserving and augmenting the backbone model's intrinsic problem-solving proficiencies.

\section{Training}

The WeMM-Video family is trained through a unified pipeline comprising progressive multimodal pre-training and multi-stage post-training.

\subsection{Pre-training}


The pre-training of WeMM-Video follows a progressive three-stage curriculum that systematically expands the context window while enriching cross-modal representations. This strategy smoothly bridges modality gaps, injects broad omni-modal knowledge, and scales long-horizon contextual comprehension, as outlined below.


\paragraph{Stage 1: Acoustic Modality Alignment.}
The initial stage focuses on bridging the acoustic representation space with the pre-trained backbone under a compact context window. In this stage, only the audio aligner (i.e., the window-level Q-Former) is optimized, while the backbone LLM and modality encoders remain frozen. Training on foundational audio-text pairs equips the model with core auditory perception capabilities, encompassing ASR and general audio captioning.


\paragraph{Stage 2: Omni-Modal Knowledge Injection.}
With an expanded context window, Stage 2 serves as the core pre-training phase for large-scale omni-modal knowledge injection. Starting from this stage, we unfreeze all model parameters except the visual encoder, as the inherited visual encoder already provides sufficiently strong visual representations. By training on a diverse mixture of textual, visual, auditory, and joint audio-visual data, this stage establishes a unified cross-modal representation space and prepares the model for native omni-modal video comprehension.


\paragraph{Stage 3: Long-Context Capability Scaling.}
In the final pre-training stage, we maintain the same trainable parameter configuration as Stage 2 while further scaling the context window to accommodate extended multimodal sequences. Training on lengthy multimodal inputs, such as long-form videos, extended audio recordings, and multi-video scenarios, enhances the model's capacity to capture long-range temporal dependencies across modalities while preserving robust performance on general downstream tasks.

\subsection{Post-training}
Following large-scale pre-training, the models acquire strong foundational omni-modal capabilities but still face several challenges. First, general language and vision-language capabilities can be further improved, particularly in complex perception and reasoning tasks. Second, generation stability remains a challenge in complex multimodal scenarios, where models may produce excessively long reasoning traces or repetitive outputs. Third, specialized capabilities require further enhancement, especially in challenging tasks such as audio-visual speaker identification, video-soundtrack matching, structured music understanding, and complex audio-visual reasoning. To address these challenges, we develop a progressive three-stage post-training pipeline consisting of multi-teacher OPD, anti-repetition MrDPO, and specialist RL, targeting general capability enhancement, generation stabilization, and specialized capability improvement, respectively.


\paragraph{Multi-Teacher OPD.}
We adopt a multi-teacher distillation strategy to enhance general language and vision-language capabilities while preserving omni-modal understanding. Specifically, stronger external teachers provide supervision for text and vision-language tasks to improve general perception and reasoning, while teachers from the same model family are employed to retain existing audio and audio-visual capabilities. We further leverage cross-modal distillation to transfer knowledge from text to audio tasks, enhancing speech understanding and instruction-following. This strategy enables complementary knowledge transfer across modalities while mitigating capability degradation.


\paragraph{Anti-Repetition MrDPO.}
We observe that the base models have a non-trivial tendency toward repetitive, endless decoding, which persists after pre-training and OPD and hinders subsequent RL training. To address this, we apply MrDPO \cite{tang2025video} to iteratively optimize the model using offline-mined hard-negative samples exhibiting repetitive decoding across video, audio, and vision-language tasks. Compared with online optimization methods, offline mining more efficiently captures these relatively rare degeneration cases. To maintain stability across multiple optimization rounds, we only update LoRA \cite{hu2021lora} adapters during each round, merging and reinitializing them between successive rounds.

\paragraph{Specialist RL.}
In the final stage, we employ reinforcement learning to enhance challenging downstream capabilities. We curate diverse and challenging training instances through offline inference and evaluation, filtering out samples that the model already solves reliably. Depending on task verifiability, we combine rule-based rewards for deterministic tasks with rubric-based and teacher-consistency rewards for open-ended tasks. A dynamic reward router assigns task-specific rewards, while instruct and thinking modes are jointly optimized through stochastic sampling. We further extend RL to long-context training with increased video frame coverage and sequence lengths, enabling the model to better adapt to extended multimodal inputs and maintain generation stability.

\section{Evaluation}

We conduct comprehensive evaluations of the WeMM-Video family across four core modalities: video, audio, image, and text, covering all four parameter scales: 2B, 4B, 9B, and 35B-A3B. 

For omni-modal video understanding, we compare WeMM-Video against leading proprietary models, including Gemini-3.0-Flash (medium thinking), Seed2.0-Mini-0428 (medium thinking), and Qwen3.5-Omni-Flash, alongside open-source omni-modal models such as Qwen3-Omni-30B-A3B-Instruct and MiniCPM-o-4.5-9B. For silent video understanding, we further include Qwen3.5-9B and Qwen3.6-35B-A3B as baselines. Across both video evaluation suites, we report both instruct and thinking mode results for WeMM-Video-35B-A3B to provide a direct comparison with non-thinking baselines and highlight reasoning gains, while reporting only the thinking mode for the 2B, 4B, and 9B variants for conciseness.

For audio understanding, we compare against Qwen3.5-Omni-Flash, Qwen3-Omni-30B-A3B-Instruct, and MiniCPM-o-4.5-9B, evaluating all WeMM-Video variants in instruct mode given the direct perceptual nature of most standard audio benchmarks. For image and text tasks, we compare against Qwen3.5-Omni-Flash, Qwen3.5-9B, and Qwen3.6-35B-A3B, and report thinking mode results across all WeMM-Video variants.

\subsection{Omni-Modal Video Evaluation}
Table~\ref{tab:av_res_35b} and Table~\ref{tab:av_res_9b} summarize the performance on joint audio-visual benchmarks. The evaluation is organized around five critical dimensions: general audio-visual understanding, audio-visual hallucination, video-music understanding, audio-visual interaction, and detailed audio-visual captioning. This evaluation suite incorporates 33 representative benchmarks, including Video-MME (with audio) \cite{fu2025video}, Video-MME-v2 (with audio) \cite{fu2026video}, Daily-Omni \cite{zhou2025daily}, WorldSense \cite{hong2026worldsense}, AVUT Human \cite{yang2025audio}, AV-SpeakerBench \cite{nguyen2026see}, Video-Holmes \cite{cheng2026video}, OmniVideoBench \cite{li2026omnivideobench}, RivaBench \cite{sun2025videosalmonno}, LVOmniBench \cite{tao2026lvomnibench}, FutureOmni \cite{chen2026futureomni}, JointAVBench \cite{chao2026jointavbench}, SciVideoBench \cite{deng2025scivideobench}, MMOU \cite{goel2026mmou}, OmniCompass-AV \cite{liu2026omnicompass}, EvaOmni \cite{hu2026evaomni}, OmniReasoningBench \cite{lin2026omnireasoning},
AVHBench Judgement \cite{sung2025avhbench}, SVHalluc \cite{zhang2026svhalluc}, MUSIC-AVQA \cite{Li2022Learning}, HarmonySet \cite{zhou2025harmonyset}, Qualcomm IVD \cite{pourreza2026can}, OmniVChat \cite{he2026omnivchat}, ODU-Bench \cite{chen2026omni}, O-IN-Bench \cite{xinOmniModalSpokenIntentions}, SocialOmni \cite{xie2026socialomni},
Omni-Cloze \cite{ma2026omni}, video-SALMONN 2 testset \cite{tang2025video}, UGCVideoCap \cite{wu2025ugc}, DVD-Bench (En) \cite{tang2026d}, OmniCap-IF \cite{wang2026omnicap}, Caption-Based QA on DailyOmni and WorldSense \cite{chen2026avocado}.

As shown in Table~\ref{tab:av_res_35b} and Table~\ref{tab:av_res_9b}, our flagship model, WeMM-Video-35B-A3B, establishes a new open-source state of the art in omni-modal video understanding, demonstrating dominance among open-source models while holding a clear advantage over proprietary models such as Seed2.0-Mini-0428 and Qwen3.5-Omni-Flash, and approaching Gemini-3.0-Flash. In general audio-visual understanding, WeMM-Video-35B-A3B (Thinking) surpasses Seed2.0-Mini-0428 on 16 out of 17 benchmarks and also outperforms Qwen3.5-Omni-Flash on 16 out of 17 benchmarks. Beyond general comprehension, WeMM-Video-35B-A3B also performs on par with leading proprietary models across specialized dimensions, including audio-visual hallucination, video-music understanding, audio-visual interaction, and audio-visual captioning. Moreover, WeMM-Video stands out among existing open-source models for its strong native omni-modal thinking capability: as evidenced by comparing the instruct and thinking modes of WeMM-Video-35B-A3B, explicit step-by-step reasoning unlocks substantial performance gains across most audio-visual comprehension and complex reasoning tasks.

Across the smaller variants, the WeMM-Video family similarly delivers strong performance. Notably, WeMM-Video-9B (Thinking) outperforms the leading open-source omni-modal models, Qwen3-Omni-30B-A3B-Instruct and MiniCPM-o-4.5-9B, across most general audio-visual understanding and downstream tasks, while WeMM-Video-4B (Thinking) also surpasses the 30B-scale baseline on the vast majority of benchmarks. Even our most compact WeMM-Video-2B model matches or outperforms the 30B-scale baseline on numerous general perception, hallucination, and captioning benchmarks, offering a highly competitive choice for resource-constrained deployments.

\begin{table}[!tp]
\centering
\caption{Comparison of WeMM-Video-35B-A3B with leading proprietary omni-modal models on joint audio-visual video benchmarks. Best results excluding the gray-shaded reference column are in bold.}
\label{tab:av_res_35b}
\resizebox{\columnwidth}{!}{%
\begin{tabular}{l>{\columncolor{black!5}\color{black!70}}ccccc}
\toprule
\multirow{2}{*}{\textbf{Benchmark}\textsuperscript{\ref{note:t4:priority}}} &
\textbf{Gemini-3.0-} & \textbf{Seed2.0-Mini-} & \textbf{Qwen3.5-Omni-} &
\multicolumn{2}{c}{\textbf{WeMM-Video-35B-A3B}} \\
& \textbf{Flash} & \textbf{0428} & \textbf{Flash} & \textbf{(Instruct)} & \textbf{(Thinking)} \\
\midrule
\multicolumn{6}{c}{\textit{General Audio-Visual Understanding}} \\
\midrule
Video-MME$_\text{w/ audio}$ & 88.6 & 83.6 & 79.3 & 84.2 & {\bfseries 86.4} \\
Video-MME-v2$_\text{w/ audio}$ & 63.2 & 51.7 & 36.6 & 47.5 & {\bfseries 54.9} \\
Daily-Omni & 82.4 & 82.5 & 81.8 & 81.5 & {\bfseries 83.4} \\
WorldSense & 63.8 & 61.3 & 57.9 & 59.6 & {\bfseries 62.7} \\
AVUT Human & 86.4 & 83.0 & 81.4 & 83.6 & {\bfseries 85.9} \\
AV-SpeakerBench & 75.9 & 68.5 & 65.2 & 66.5 & {\bfseries 72.6} \\
Video-Holmes & 71.2 & 62.1 & 61.7 & 63.2 & {\bfseries 63.4} \\
OmniVideoBench & 63.3 & 51.3 & 43.7 & 48.6 & {\bfseries 55.3} \\
RivaBench & 72.3 & 68.3 & 68.5 & 67.9 & {\bfseries 70.8} \\
LVOmniBench & 59.8 & 49.7 & 43.7 & 50.3 & {\bfseries 52.6} \\
FutureOmni & 65.3 & 52.2 & 55.4 & {\bfseries 58.4} & 56.2 \\
JointAVBench & 68.4 & 64.7 & {\bfseries 69.0} & 67.4 & 65.9 \\
SciVideoBench & 75.2 & 60.1 & 51.3 & 57.8 & {\bfseries 69.4} \\
MMOU (test-mini) & 84.7 & 65.8 & 69.9 & 71.4 & {\bfseries 74.5} \\
OmniCompass-AV & 50.8 & {\bfseries 47.9} & 39.5 & 41.5 & 46.3 \\
EvaOmni & 66.4 & 54.3 & 30.4 & 48.8 & {\bfseries 57.8} \\
OmniReasoningBench & 52.4 & 42.9 & 37.2 & 44.3 & {\bfseries 45.4} \\
\midrule
\multicolumn{6}{c}{\textit{Audio-Visual Hallucination}} \\
\midrule
AVHBench Judgement\textsuperscript{\ref{note:t4:avhbench}} & 78.5 & 75.8 & 77.5 & {\bfseries 80.3} & 80.0 \\
SVHalluc\textsuperscript{\ref{note:t4:svhalluc}} & 86.4 & 75.4 & 63.6 & 75.3 & {\bfseries 80.3} \\
\midrule
\multicolumn{6}{c}{\textit{Video-Music Understanding}} \\
\midrule
MUSIC-AVQA & 76.2 & 74.2 & {\bfseries 82.1} & 79.6 & 79.5 \\
HarmonySet\textsuperscript{\ref{note:t4:harmonyset}} & 5.73 & 5.74 & 5.48 & 5.63 & {\bfseries 5.97} \\
\midrule
\multicolumn{6}{c}{\textit{Audio-Visual Interaction}\textsuperscript{\ref{note:t4:avinteract}}} \\
\midrule
Qualcomm IVD & 72.2 & 70.0 & 66.3 & 70.2 & {\bfseries 72.3} \\
OmniVChat & 54.5 & {\bfseries 49.8} & 36.9 & 44.8 & 48.9 \\
ODU-Bench\textsuperscript{\ref{note:t4:odubench}} & 72.2 \textbar{} 72.5 & 69.8 \textbar{} 73.6 & 64.4 \textbar{} 70.4 & {\bfseries 74.2} \textbar{} {\bfseries 78.9} & 72.5 \textbar{} 77.8 \\
O-IN-Bench & 53.3 & {\bfseries 49.4} & 34.2 & 38.4 & 49.0 \\
SocialOmni\textsuperscript{\ref{note:t4:socialomni}} & 90.2 \textbar{} 79.5 \textbar{} 87.9 & 89.1 \textbar{} 49.0 \textbar{} {\bfseries 90.0} & 87.4 \textbar{} {\bfseries 67.0} \textbar{} 43.6 & {\bfseries 90.6} \textbar{} 43.5 \textbar{} 74.8 & 90.5 \textbar{} 62.0 \textbar{} 85.2 \\
\midrule
\multicolumn{6}{c}{\textit{Audio-Visual Captioning}\textsuperscript{\ref{note:t4:avcaption}}} \\
\midrule
Omni-Cloze & 34.3 & 39.9 & 41.8 & {\bfseries 44.2} & 42.6 \\
video-SALMONN 2 testset\textsuperscript{\ref{note:t4:vs2}} & 37.7 & 32.7 & 28.0 & {\bfseries 27.5} & 28.3 \\
UGCVideoCap & 71.0 & 74.8 & 70.5 & {\bfseries 75.1} & 74.9 \\
DVD-Bench (En)\textsuperscript{\ref{note:t4:dvdbench}} & 81.2 & 75.3 & 72.5 & 76.2 & {\bfseries 79.1} \\
OmniCap-IF\textsuperscript{\ref{note:t4:omnicap}} & 83.2 \textbar{} 31.1 & 75.1 \textbar{} 16.8 & 65.6 \textbar{} 9.2 & 74.8 \textbar{} 16.9 & {\bfseries 79.1} \textbar{} {\bfseries 23.6} \\
DailyOmni (Caption-Based QA) & 51.8 & 50.9 & 44.0 & 51.9 & {\bfseries 62.2} \\
WorldSense (Caption-Based QA) & 32.4 & 28.5 & 26.0 & 32.3 & {\bfseries 36.6} \\
\bottomrule
\end{tabular}%
}

\vspace{0.2cm}
\begin{minipage}{\columnwidth}
\footnotesize
\begin{enumerate}[label=\textsuperscript{\arabic*}, ref=\arabic*,
                  leftmargin=1.5em, labelsep=0.4em, nosep]
\item \label{note:t4:priority}
For existing models, we report official results where available and our own evaluations otherwise.
\item \label{note:t4:avhbench}
AVHBench results are reported as the sample-wise accuracy across all samples in the judgment tasks.
\item \label{note:t4:svhalluc}
SVHalluc results are reported as the sample-wise accuracy across all samples.
\item \label{note:t4:harmonyset}
HarmonySet results are reported as the overall score judged by GPT-4o following official settings.
\item \label{note:t4:avinteract}
In audio-visual interaction evaluation, O-IN-Bench is evaluated with GPT-6-Luna as the judge following official settings, and the other benchmarks with GPT-6.1-Sol.
\item \label{note:t4:odubench}
ODU-Bench results are reported as the average score of the AV \textbar{} AO subsets, respectively.
\item \label{note:t4:socialomni}
SocialOmni results are reported in the order Who \textbar{} When \textbar{} How (QGold).
\item \label{note:t4:avcaption}
In audio-visual captioning evaluation, Omni-Cloze and UGCVideoCap results are judged by GPT-4o following official settings, DVD-Bench (En) results are judged by Qwen3-235B-A22B following official settings, and results of other benchmarks are judged by GPT-6.1-Sol.
\item \label{note:t4:vs2}
video-SALMONN 2 testset results are reported as the total error rate, where lower values indicate better performance.
\item \label{note:t4:dvdbench}
DVD-Bench (En) results are reported as speaker accuracy.
\item \label{note:t4:omnicap}
OmniCap-IF results are reported as CSR \textbar{} ISR.
\end{enumerate}
\end{minipage}
\end{table}

\begin{table}[!tp]
\centering
\caption{Comparison of WeMM-Video-2B, 4B, and 9B with open-source omni-modal baselines on joint audio-visual video benchmarks. Best results are in bold.}
\label{tab:av_res_9b}
\resizebox{\columnwidth}{!}{%
\begin{tabular}{lccccc}
\toprule
\multirow{2}{*}{\textbf{Benchmark}\textsuperscript{\ref{note:av-res-9b:priority}}} &
\textbf{Qwen3-Omni-} & \textbf{MiniCPM-o} &
\multicolumn{3}{c}{\textbf{WeMM-Video-}} \\
& \textbf{30B-A3B-Instruct} & \textbf{4.5-9B} & \textbf{2B (Thinking)} &
\textbf{4B (Thinking)} & \textbf{9B (Thinking)} \\
\midrule
\multicolumn{6}{c}{\textit{General Audio-Visual Understanding}} \\
\midrule
Video-MME$_\text{w/ audio}$ & 77.3 & 69.2 & 72.4 & 83.9 & {\bfseries 85.1} \\
Video-MME-v2$_\text{w/ audio}$ & 36.3 & 28.1 & 37.7 & 48.1 & {\bfseries 51.6} \\
Daily-Omni & 70.7 & 80.2 & 76.5 & 82.5 & {\bfseries 83.6} \\
WorldSense & 54.0 & 55.7 & 52.4 & 60.3 & {\bfseries 63.4} \\
AVUT Human & 74.2 & 78.6 & 80.2 & 84.3 & {\bfseries 85.8} \\
AV-SpeakerBench & 50.4 & 56.7 & 56.2 & 68.6 & {\bfseries 71.1} \\
Video-Holmes & 57.6 & {\bfseries 64.3} & 48.8 & 58.8 & 61.3 \\
OmniVideoBench & 45.4 & 37.2 & 39.2 & 49.7 & {\bfseries 52.8} \\
RivaBench & 69.6 & 68.7 & 63.8 & 69.6 & {\bfseries 70.1} \\
LVOmniBench & 42.7 & 37.0 & 40.2 & 48.1 & {\bfseries 51.0} \\
FutureOmni & 55.1 & {\bfseries 56.1} & 45.2 & 50.9 & 53.7 \\
JointAVBench &  64.6 & 60.0 & 60.3 & {\bfseries 65.1} & 64.6 \\
SciVideoBench & 49.9 & 32.7 & 39.6 & 59.2 & {\bfseries 64.2} \\
MMOU (test-mini) & 60.5 & 40.9 & 45.7 & 63.3 & {\bfseries 71.7} \\
OmniCompass-AV & 28.7 & 21.5 & 29.1 & 40.7 & {\bfseries 42.7} \\
EvaOmni & 26.1 & 21.2 & 32.8 & 46.5 & {\bfseries 50.7} \\
OmniReasoningBench & 24.1 & 16.3 & 27.6 & 39.5 & {\bfseries 43.9} \\
\midrule
\multicolumn{6}{c}{\textit{Audio-Visual Hallucination}} \\
\midrule
AVHBench Judgement\textsuperscript{\ref{note:av-res-9b:avhbench}} & 74.5 & 79.2 & 74.3 & 75.4 & {\bfseries 80.0} \\
SVHalluc\textsuperscript{\ref{note:av-res-9b:svhalluc}} & 65.2 & 68.9 & 73.9 & 80.2 & {\bfseries 86.8} \\
\midrule
\multicolumn{6}{c}{\textit{Video-Music Understanding}} \\
\midrule
MUSIC-AVQA & 71.0 & 71.6 & 75.4 & 78.0 & {\bfseries 79.3} \\
HarmonySet\textsuperscript{\ref{note:av-res-9b:harmonyset}} & 5.15 & 5.55 & {\bfseries 6.02} & 5.94 & 5.98 \\
\midrule
\multicolumn{6}{c}{\textit{Audio-Visual Interaction}\textsuperscript{\ref{note:av-res-9b:avinteract}}} \\
\midrule
Qualcomm IVD & 64.6 & 43.0 & 62.3 & 69.9 & {\bfseries 70.3} \\
OmniVChat & 46.5 & 30.4 & 33.1 & 44.3 & {\bfseries 49.5} \\
ODU-Bench\textsuperscript{\ref{note:av-res-9b:odubench}} & 57.9 \textbar{} 65.5 & 47.8 \textbar{} 58.4 & 56.5 \textbar{} 62.1 & 70.2 \textbar{} 73.1 & {\bfseries 72.6} \textbar{} {\bfseries 75.6} \\
O-IN-Bench & 35.7 & 29.3 & 32.4 & 42.0 & {\bfseries 44.1} \\
SocialOmni\textsuperscript{\ref{note:av-res-9b:socialomni}} & 74.7 \textbar{} 58.0 \textbar{} 44.5 & 83.9 \textbar{} 39.5 \textbar{} 67.4 & 82.0 \textbar{} 41.0 \textbar{} 44.5 & {\bfseries 90.9} \textbar{} 57.5 \textbar{} 80.3 & 90.6 \textbar{} {\bfseries 59.5} \textbar{} {\bfseries 84.4} \\
\midrule
\multicolumn{6}{c}{\textit{Audio-Visual Captioning}\textsuperscript{\ref{note:av-res-9b:avcaption}}} \\
\midrule
Omni-Cloze & 27.0 & 27.1 & 38.4 & 40.1 & {\bfseries 41.5} \\
video-SALMONN 2 testset\textsuperscript{\ref{note:av-res-9b:vs2}} & {\bfseries 27.9} & 50.6 & 34.3 & 30.4 & 30.9 \\
UGCVideoCap & 70.8 & 56.9 & 72.2 & {\bfseries 75.1} & 74.4 \\
DVD-Bench (En)\textsuperscript{\ref{note:av-res-9b:dvdbench}} & 67.8 & 64.9 & 73.6 & {\bfseries 78.5} & 78.3 \\
OmniCap-IF\textsuperscript{\ref{note:av-res-9b:omnicap}} & 62.7 \textbar{} 7.2 & 64.7 \textbar{} 9.3 & 63.3 \textbar{} 6.8 & 73.2 \textbar{} 14.0 & {\bfseries 76.7} \textbar{} {\bfseries 20.0} \\
DailyOmni (Caption-Based QA) & 19.4 & 26.6 & 51.1 & 58.6 & {\bfseries 59.7} \\
WorldSense (Caption-Based QA) & 14.6 & 16.6 & 29.6 & {\bfseries 34.6} & 34.1 \\
\bottomrule
\end{tabular}%
}

\vspace{0.2cm}
\begin{minipage}{\columnwidth}
\footnotesize
\begin{enumerate}[label=\textsuperscript{\arabic*}, ref=\arabic*,
                  leftmargin=1.5em, labelsep=0.4em, nosep]
\item \label{note:av-res-9b:priority}
For existing models, we report official results where available and our own evaluations otherwise.
\item \label{note:av-res-9b:avhbench}
AVHBench results are reported as the sample-wise accuracy across all samples in the judgment tasks.
\item \label{note:av-res-9b:svhalluc}
SVHalluc results are reported as the sample-wise accuracy across all samples.
\item \label{note:av-res-9b:harmonyset}
HarmonySet results are reported as the overall score judged by GPT-4o following official settings.
\item \label{note:av-res-9b:avinteract}
In audio-visual interaction evaluation, O-IN-Bench is evaluated with GPT-6-Luna as the judge following official settings, and the other benchmarks with GPT-6.1-Sol.
\item \label{note:av-res-9b:odubench}
ODU-Bench results are reported as the average score of the AV \textbar{} AO subsets, respectively.
\item \label{note:av-res-9b:socialomni}
SocialOmni results are reported in the order Who \textbar{} When \textbar{} How (QGold).
\item \label{note:av-res-9b:avcaption}
In audio-visual captioning evaluation, Omni-Cloze and UGCVideoCap results are judged by GPT-4o following official settings, DVD-Bench (En) results are judged by Qwen3-235B-A22B following official settings, and results of other benchmarks are judged by GPT-6.1-Sol.
\item \label{note:av-res-9b:vs2}
video-SALMONN 2 testset results are reported as the total error rate, where lower values indicate better performance.
\item \label{note:av-res-9b:dvdbench}
DVD-Bench (En) results are reported as speaker accuracy.
\item \label{note:av-res-9b:omnicap}
OmniCap-IF results are reported as CSR \textbar{} ISR.
\end{enumerate}
\end{minipage}
\end{table}

\subsection{Silent Video Evaluation}

\begin{table}[!tp]
\centering
\caption{Comparison of WeMM-Video-35B-A3B with proprietary and open-source baselines on silent video benchmarks. Best results excluding the gray-shaded reference column are in bold.}
\label{tab:vd_res_35b}
\resizebox{\columnwidth}{!}{%
\begin{tabular}{l>{\columncolor{black!5}\color{black!70}}cccccc}
\toprule
\multirow{2}{*}{\textbf{Benchmark}\textsuperscript{\ref{note:t2:priority}}} &
\textbf{Gemini-3.0-} & \textbf{Seed2.0-Mini-} & \textbf{Qwen3.5-Omni-} & \textbf{Qwen3.6-} &
\multicolumn{2}{c}{\textbf{WeMM-Video-35B-A3B}} \\
& \textbf{Flash} & \textbf{0428} & \textbf{Flash} & \textbf{35B-A3B} & \textbf{(Instruct)} & \textbf{(Thinking)} \\
\midrule
\multicolumn{7}{c}{\textit{General Video Understanding}} \\
\midrule
Video-MME & 86.4 & 78.1 & 77.0 & {\bfseries 82.5} & 78.0 & 80.4 \\
Video-MME-v2 & 55.6 & 44.4 & 33.7 & {\bfseries 45.3} & 39.8 & 45.2 \\
MVBench & 70.0 & 73.4 & 70.8 & {\bfseries 74.6} & 72.8 & 73.1 \\
Video-MMMU & 86.9 & 81.0 & 64.0 & {\bfseries 83.7} & 80.8 & 83.6 \\
MME-VideoOCR & 81.5 & {\bfseries 80.5} & 70.5 & 75.8 & 73.3 & 78.4 \\
TempCompass & 88.1 & {\bfseries 85.9} & 78.1 & 83.7 & 82.4 & 84.3 \\
NExT-QA & 83.0 & 83.9 & 84.1 & 84.3 & {\bfseries 85.7} & 84.6 \\
TOMATO & 45.9 & {\bfseries 53.8} & 28.2 & 51.4 & 46.4 & 52.5 \\
Video-IFBench\textsuperscript{\ref{note:t2:ifbench}} & 72.2 \textbar{} 49.5 & 56.5 \textbar{} 28.9 & 53.6 \textbar{} 26.6 & 51.2 \textbar{} {\bfseries 34.1} & 50.8 \textbar{} 24.7 & {\bfseries 57.8} \textbar{} 31.8 \\
VidHalluc\textsuperscript{\ref{note:t2:vidhalluc}} & 80.0 & 76.7 & 76.1 & 84.0 & 80.2 & {\bfseries 84.1} \\
VideoVista & 83.2 & 81.8 & 84.1 & 83.1 & {\bfseries 84.5} & 83.5 \\
MMVU & 77.1 & 70.5 & 62.7 & {\bfseries 74.7} & 68.5 & 74.4 \\
Video-Index & 58.8 & {\bfseries 38.3} & 24.3 & 38.1 & 27.5 & 37.9 \\
Perception Test & 78.9 & {\bfseries 81.1} & 74.1 & 79.8 & 77.9 & 80.4 \\
TimeBlind & 77.9 & {\bfseries 80.6} & 67.8 & 77.0 & 73.8 & 78.0 \\
Video-MME-Logical & 31.5 & {\bfseries 28.9} & 19.7 & 27.5 & 25.1 & 26.0 \\

\midrule
\multicolumn{7}{c}{\textit{Long Video Understanding}} \\
\midrule
MLVU & 87.9 & 81.6 & 81.9 & {\bfseries 86.2} & 80.8 & 85.3 \\
LVBench & 73.8 & 63.4 & 65.7 & 71.4 & 68.7 & {\bfseries 71.6} \\
LongVideoBench & 77.4 & 74.4 & 66.9 & {\bfseries 76.7} & 74.0 & 76.6 \\
EgoSchema & 64.8 & {\bfseries 78.4} & 70.0 & 77.2 & 76.0 & 74.2 \\
MMBench-Video\textsuperscript{\ref{note:t2:mmbench}} & 2.56 & 2.48 & 2.06 & 2.49 & 2.43 & {\bfseries 2.50} \\

\midrule
\multicolumn{7}{c}{\textit{Fine-grained Perception}} \\
\midrule
MotionBench & 65.0 & {\bfseries 68.4} & 61.7 & 66.1 & 67.1 & 65.7 \\
TemporalBench\textsuperscript{\ref{note:t2:temporalbench}} & 50.9 & 44.9 & 30.6 & 43.3 & 39.0 & {\bfseries 45.9} \\
F-16-NBA\textsuperscript{\ref{note:t2:f16nba}} & 90.1 & {\bfseries 94.8} & 87.3 & 82.8 & 89.0 & 89.9 \\

\midrule
\multicolumn{7}{c}{\textit{Temporal Grounding}} \\
\midrule
TimeLens-Charades\textsuperscript{\ref{note:t2:timelens}} & 58.8 & {\bfseries 56.9} & 40.8 & 47.4 & 54.7 & 54.8 \\
TimeLens-ActivityNet & 47.1 & 56.2 & 30.3 & 53.9 & {\bfseries 58.9} & 56.4 \\
TimeLens-QVHighlights & 68.2 & 68.5 & 37.3 & 66.8 & 68.4 & {\bfseries 69.0} \\

\midrule
\multicolumn{7}{c}{\textit{Spatial Understanding}} \\
\midrule
PAI-Bench-U & 65.8 & {\bfseries 67.7} & 51.5 & 65.3 & 62.5 & 65.5 \\
MMSI-Video-Bench & 43.2 & 34.1 & 29.5 & 32.7 & 31.1 & {\bfseries 35.6} \\
VSI-Bench & 53.3 & 49.4 & 55.2 & 56.1 & {\bfseries 62.7} & 57.2 \\

\midrule
\multicolumn{7}{c}{\textit{Streaming Video Understanding}} \\
\midrule
OVO-Bench & 69.3 & {\bfseries 70.6} & 55.7 & 61.0 & 63.1 & 63.9 \\

\midrule
\multicolumn{7}{c}{\textit{Multi-Video Understanding}} \\
\midrule
CrossVid & 53.2 & 49.9 & 40.4 & 45.2 & 45.0 & {\bfseries 50.2} \\

\bottomrule
\end{tabular}%
}

\vspace{0.2cm}
\begin{minipage}{\columnwidth}
\footnotesize
\begin{enumerate}[
  label=\textsuperscript{\arabic*},
  ref=\arabic*,
  leftmargin=1.5em,
  labelsep=0.4em,
  nosep
]
\item \label{note:t2:priority}
For existing models, we report official results where available and our own evaluations otherwise.

\item \label{note:t2:ifbench}
Video-IFBench results are reported as TCSR \textbar{} TISR.

\item \label{note:t2:vidhalluc}
VidHalluc results are reported as the average score across the four subtasks: BQA, MCQ, TSH, and STH.

\item \label{note:t2:mmbench}
MMBench-Video scores are reported on a scale of 0--3.

\item \label{note:t2:temporalbench}
TemporalBench results are reported using the overall MBA score.

\item \label{note:t2:f16nba}
F-16-NBA is reported as accuracy.

\item \label{note:t2:timelens}
TimeLens results are reported as mean IoU.

\end{enumerate}
\end{minipage}

\end{table}

\begin{table}[!tp]
\centering
\caption{Comparison of WeMM-Video-2B, 4B, and 9B with open-source baselines on silent video benchmarks. Best results are in bold.}
\label{tab:vd_res_9b}
\resizebox{\columnwidth}{!}{%
\begin{tabular}{lcccccc}
\toprule
\multirow{2}{*}{\textbf{Benchmark}\textsuperscript{\ref{note:vd-res-9b:priority}}} &
\textbf{Qwen3-Omni-} & \textbf{Qwen3.5-} & \textbf{MiniCPM-o-} &
\multicolumn{3}{c}{\textbf{WeMM-Video-}} \\
& \textbf{30B-A3B-Instruct} & \textbf{9B} & \textbf{4.5-9B} &
\textbf{2B (Thinking)} & \textbf{4B (Thinking)} & \textbf{9B (Thinking)} \\
\midrule
\multicolumn{7}{c}{\textit{General Video Understanding}} \\
\midrule
Video-MME & 70.5 & 78.4 & 70.4 & 67.6 & 75.7 & {\bfseries 78.5} \\
Video-MME-v2 & 29.7 & {\bfseries 42.4} & 26.8 & 32.8 & 39.5 & 40.6 \\
MVBench & 68.8 & {\bfseries 74.4} & 63.4 & 64.6 & 70.0 & 71.4 \\
Video-MMMU & 54.7 & 78.9 & 52.9 & 65.1 & {\bfseries 79.6} & 79.2 \\
MME-VideoOCR & 67.7 & 75.9 & 55.4 & 67.7 & 75.0 & {\bfseries 77.3} \\
TempCompass & 76.5 & 82.8 & 74.4 & 78.5 & {\bfseries 84.9} & 84.8 \\
NExT-QA & {\bfseries 84.8} & 81.9 & 80.2 & 76.6 & 81.6 & 83.1 \\
TOMATO & 35.2 & 45.0 & 29.4 & 34.4 & 47.9 & {\bfseries 48.5} \\
Video-IFBench\textsuperscript{\ref{note:vd-res-9b:ifbench}} & 44.3 \textbar{} 18.0 & {\bfseries 56.0} \textbar{} {\bfseries 32.2} & 23.2 \textbar{} 10.4 & 33.1 \textbar{} 13.5 & 48.5 \textbar{} 22.2 & 52.8 \textbar{} 26.1 \\
VidHalluc\textsuperscript{\ref{note:vd-res-9b:vidhalluc}} & 76.2 & 81.9 & 75.2 & 77.5 & 82.5 & {\bfseries 83.0} \\
VideoVista & 83.7 & {\bfseries 84.7} & 82.5 & 78.8 & 81.2 & 82.2 \\
MMVU & 62.9 & 69.6 & 59.2 & 50.8 & 64.9 & {\bfseries 70.2} \\
Video-Index & 19.6 & 34.2 & 11.7 & 18.3 & 30.5 & {\bfseries 34.6} \\
Perception Test & 75.3 & 77.5 & 67.8 & 71.2 & 76.5 & {\bfseries 78.0} \\
TimeBlind & 66.6 & {\bfseries 76.6} & 60.5 & 67.5 & 75.2 & 76.4 \\
Video-MME-Logical & 15.9 & {\bfseries 24.1} & 14.7 & 18.7 & 21.7 & 21.8 \\

\midrule
\multicolumn{7}{c}{\textit{Long Video Understanding}} \\
\midrule
MLVU & 75.2 & {\bfseries 84.4} & 76.5 & 74.4 & 80.4 & 82.4 \\
LVBench & 50.2 & {\bfseries 70.0} & 50.9 & 55.8 & 65.3 & 68.2 \\
LongVideoBench & 63.5 & {\bfseries 74.8} & 66.0 & 66.0 & {\bfseries 74.8} & 74.7 \\
EgoSchema & 66.4 & {\bfseries 74.4} & 60.4 & 66.4 & 74.2 & 73.6 \\
MMBench-Video\textsuperscript{\ref{note:vd-res-9b:mmbench}} & 2.13 & 2.16 & 2.20 & 2.20 & 2.38 & {\bfseries 2.44} \\

\midrule
\multicolumn{7}{c}{\textit{Fine-grained Perception}} \\
\midrule
MotionBench & 61.9 & 63.3 & 61.4 & 58.0 & 63.4 & {\bfseries 64.0} \\
TemporalBench\textsuperscript{\ref{note:vd-res-9b:temporalbench}} & 34.7 & 39.5 & 23.9 & 29.0 & 38.3 & {\bfseries 40.9} \\
F-16-NBA\textsuperscript{\ref{note:vd-res-9b:f16nba}} & 74.0 & 75.8 & 77.1 & 61.4 & 87.0 & {\bfseries 88.7} \\

\midrule
\multicolumn{7}{c}{\textit{Temporal Grounding}} \\
\midrule
TimeLens-Charades\textsuperscript{\ref{note:vd-res-9b:timelens}} & 45.5 & 50.6 & 29.0 & 49.7 & {\bfseries 54.9} & 52.3 \\
TimeLens-ActivityNet & 38.2 & 52.2 & 15.8 & 43.3 & {\bfseries 55.3} & {\bfseries 55.3} \\
TimeLens-QVHighlights & 47.5 & 64.3 & 12.9 & 48.7 & 66.0 & {\bfseries 66.8} \\

\midrule
\multicolumn{7}{c}{\textit{Spatial Understanding}} \\
\midrule
PAI-Bench-U & 59.2 & 63.4 & 48.4 & 52.9 & 62.7 & {\bfseries 64.8} \\
MMSI-Video-Bench & 29.5 & 31.2 & 27.3 & 26.3 & 32.0 & {\bfseries 32.4} \\
VSI-Bench & 55.7 & {\bfseries 58.3} & 44.0 & 44.1 & 51.1 & 54.5 \\

\midrule
\multicolumn{7}{c}{\textit{Streaming Video Understanding}} \\
\midrule
OVO-Bench & 60.0 & 60.8 & 56.4 & 57.2 & 61.6 & {\bfseries 63.3} \\

\midrule
\multicolumn{7}{c}{\textit{Multi-Video Understanding}} \\
\midrule
CrossVid & 32.4 & 37.8 & 33.1 & 31.8 & 39.9 & {\bfseries 45.2} \\

\bottomrule
\end{tabular}%
}

\vspace{0.2cm}
\begin{minipage}{\columnwidth}
\footnotesize
\begin{enumerate}[
  label=\textsuperscript{\arabic*},
  ref=\arabic*,
  leftmargin=1.5em,
  labelsep=0.4em,
  nosep
]
\item \label{note:vd-res-9b:priority}
For existing models, we report official results where available and our own evaluations otherwise.

\item \label{note:vd-res-9b:ifbench}
Video-IFBench results are reported as TCSR \textbar{} TISR.

\item \label{note:vd-res-9b:vidhalluc}
VidHalluc results are reported as the average score across the four subtasks: BQA, MCQ, TSH, and STH.

\item \label{note:vd-res-9b:mmbench}
MMBench-Video scores are reported on a scale of 0--3.

\item \label{note:vd-res-9b:temporalbench}
TemporalBench results are reported using the overall MBA score.

\item \label{note:vd-res-9b:f16nba}
F-16-NBA is reported as accuracy.

\item \label{note:vd-res-9b:timelens}
TimeLens results are reported as mean IoU.

\end{enumerate}
\end{minipage}
\end{table}



Table~\ref{tab:vd_res_35b} and Table~\ref{tab:vd_res_9b} present the evaluation results on silent video benchmarks. We evaluate the models across seven core dimensions: general video understanding, long video understanding, fine-grained perception, temporal grounding, spatial understanding, streaming video understanding, and multi-video understanding. The evaluation suite encompasses 32 established benchmarks, including Video-MME \cite{fu2025video}, Video-MME-v2 \cite{fu2026video}, MVBench \cite{li2024mvbench}, Video-MMMU \cite{hu2026video}, MME-VideoOCR \cite{shi2026mme}, TempCompass \cite{liu2024tempcompass}, NExT-QA \cite{xiao2021next}, TOMATO \cite{shangguan2025tomato}, Video-IFBench \cite{liu2026video}, VidHalluc \cite{li2025vidhalluc}, VideoVista \cite{li2024videovista}, MMVU \cite{zhao2025mmvu}, Video-Index \cite{song2026video}, Perception Test \cite{patraucean2023perception}, TimeBlind \cite{li2026timeblind}, Video-MME-Logical \cite{kwan2026video}, MLVU \cite{zhou2025mlvu}, LVBench \cite{wang2025lvbench}, LongVideoBench \cite{wu2024longvideobench}, EgoSchema \cite{mangalam2023egoschema}, MMBench-Video \cite{fang2024mmbench}, MotionBench \cite{hong2025motionbench}, TemporalBench \cite{cai2024temporalbench}, F-16-NBA \cite{li2025improving}, TimeLens \cite{zhang2026timelens}, PAI-Bench-U \cite{zhou2026pai}, MMSI-Video-Bench \cite{lin2025mmsi}, VSI-Bench \cite{yang2025thinking}, OVO-Bench \cite{niu2025ovo}, CrossVid \cite{li2026crossvid}.

Overall, the WeMM-Video family demonstrates highly competitive performance on silent video understanding without suffering much from the visual degradation often introduced by acoustic modality integration. In particular, WeMM-Video-35B-A3B holds a clear lead over Qwen3.5-Omni-Flash across nearly all benchmarks, while performing on par with its strong vision-language backbone, Qwen3.6-35B-A3B, as well as the proprietary Seed2.0-Mini-0428 across evaluated dimensions. Consistent trends are observed across the smaller variants: WeMM-Video-9B and WeMM-Video-4B substantially outperform the open-source omni-modal baselines Qwen3-Omni-30B-A3B-Instruct and MiniCPM-o-4.5-9B, with WeMM-Video-9B matching or surpassing its Qwen3.5-9B backbone on the majority of benchmarks and WeMM-Video-2B delivering competitive results at a compact scale.

\subsection{Audio Evaluation}

Table \ref{tab:audio_res} summarizes the evaluation results on standalone audio benchmarks across five core dimensions: general audio understanding, music understanding, speech recognition, lyrics recognition, and voice interaction. The evaluation suite encompasses 18 established benchmarks, including MMAU \cite{sakshi2025mmau}, MMAU-Pro \cite{kumar2026mmau}, MMAR \cite{ma2026mmar}, MMSU \cite{wang2026mmsu}, DCase 2025 AQA (dev subset) \cite{yang2026multi},
RUL-MuchoMusic \cite{zang2025you},
HumMusQA \cite{weck2026hummusqa}, MusTBench \cite{kwon2026MusTBench},
GigaSpeech \cite{chen2021gigaspeech}, VoxPopuli \cite{wang2021voxpopuli}, LibriSpeech \cite{panayotov2015librispeech}, WenetSpeech \cite{zhang2022wenetspeech}, MIR-1K \cite{hsu2009improvement}, Opencpop \cite{wang2022opencpop}, VoiceBench \cite{chen2026voicebench}, IFEval-Audio \cite{gao2025ifeval}, Speech-IFEval \cite{lu2025speech}, 
and Audio MultiChallenge \cite{gosai2026audio}.

\begin{table}[!htbp]
\centering
\caption{Evaluation results of the WeMM-Video family and omni-modal baselines on standalone audio benchmarks. Best results are in bold.}
\label{tab:audio_res}
\resizebox{\columnwidth}{!}{%
\begin{tabular}{lccccccc}
\toprule
\multirow{2}{*}{\textbf{Benchmark}} & \textbf{Qwen3.5-Omni-} & \textbf{Qwen3-Omni-} & \textbf{MiniCPM-o} & \multicolumn{4}{c}{\textbf{WeMM-Video-}} \\
& \textbf{Flash} & \textbf{30B-A3B-Instruct} & \textbf{4.5-9B} & \textbf{2B (Instruct)} & \textbf{4B (Instruct)} & \textbf{9B (Instruct)} & \textbf{35B-A3B (Instruct)} \\
\midrule
\multicolumn{8}{c}{\textit{General Audio Understanding}} \\
\midrule
MMAU & {\bfseries 80.4} & 77.5 & 76.9 & 72.4 & 75.2 & 77.0 & 78.0 \\
MMAU-Pro & 64.8 & 60.2 & 62.7 & 62.9 & 68.1 & {\bfseries 70.9} & 70.6 \\
MMAR & {\bfseries 74.0} & 73.2 & 66.3 & 56.9 & 67.3 & 72.8 & 73.6 \\
MMSU & 72.2 & 69.0 & 67.4 & 65.6 & 72.6 & {\bfseries 76.8} & 76.0 \\
DCase 2025 AQA & {\bfseries 67.2} & 62.6 & 65.9 & 58.6 & 64.9 & 66.7 & {\bfseries 67.2} \\

\midrule
\multicolumn{8}{c}{\textit{Music Understanding}} \\
\midrule
RUL-MuchoMusic & {\bfseries 60.5} & 52.0 & 43.9 & 46.8 & 55.4 & 57.2 & 57.4 \\
HumMusQA & 69.6 & 65.9 & 56.6 & 55.9 & 66.6 & 65.3 & {\bfseries 70.0} \\
MusTBench & 42.2 & 32.4 & 24.0 & 33.9 & 41.8 & 42.0 & {\bfseries 46.6} \\ 

\midrule
\multicolumn{8}{c}{\textit{Speech Recognition}\textsuperscript{\ref{note:audio-res:asr}}} \\
\midrule
GigaSpeech & 9.2 & 8.7 & 10.1 & 9.1 & 8.6 & {\bfseries 8.5} & 8.6 \\
VoxPopuli & 5.7 & 6.4 & 8.5 & 6.7 & 5.7 & {\bfseries 5.6} & 5.7 \\
LibriSpeech\textsuperscript{\ref{note:audio-res:ls}} & 1.3 \textbar{} {\bfseries 2.4} & {\bfseries 1.2} \textbar{} 2.5 & 1.4 \textbar{} 2.8 & 2.3 \textbar{} 4.7 & 1.7 \textbar{} 3.5 & 1.5 \textbar{} 3.1 & 1.6 \textbar{} 3.4 \\
WenetSpeech\textsuperscript{\ref{note:audio-res:wenet}} & {\bfseries 4.4} \textbar{} {\bfseries 5.5} & 4.7 \textbar{} 5.9 & 5.9 \textbar{} 5.7 & 6.4 \textbar{} 7.7 & 5.8 \textbar{} 6.9 & 5.4 \textbar{} 6.8 & 5.7 \textbar{} 6.8 \\

\midrule
\multicolumn{8}{c}{\textit{Lyrics Recognition}\textsuperscript{\ref{note:audio-res:lyrics}}} \\
\midrule
MIR-1K (vocal) & {\bfseries 4.9} & 5.9 & 25.9 & 9.7 & 8.3 & 8.1 & 7.9 \\
Opencpop & {\bfseries 1.1} & 1.5 & 23.0 & 3.1 & 2.6 & 2.8 & 2.4 \\

\midrule
\multicolumn{8}{c}{\textit{Voice Interaction}} \\
\midrule
VoiceBench & 87.8 & 85.5 & 68.3 & 66.5 & 83.0 & {\bfseries 89.0} & 88.4 \\
IFEval-Audio & 58.2 & 57.9 & 52.1 & 44.3 & 56.1 & 59.6 & {\bfseries 63.6} \\
Speech-IFEval & 88.2 & 77.6 & 64.8 & 94.1 & 91.7 & 95.7 & {\bfseries 96.5} \\
Audio MultiChallenge\textsuperscript{\ref{note:audio-res:audio-mc}} & 58.2 & 52.2 & 45.4 & 37.7 & 54.9 & 59.2 & {\bfseries 62.0} \\

\bottomrule
\end{tabular}%
}

\vspace{0.2cm}
\begin{minipage}{\columnwidth}
\footnotesize
\begin{enumerate}[label=\textsuperscript{\arabic*}, ref=\arabic*,
                  leftmargin=1.5em, labelsep=0.4em, nosep]
\item \label{note:audio-res:asr}
GigaSpeech, VoxPopuli, and LibriSpeech results are reported as WER. WenetSpeech results are reported as CER.
\item \label{note:audio-res:ls}
LibriSpeech results are reported as the WER results of test-clean \textbar{} test-other subsets.
\item \label{note:audio-res:wenet}
WenetSpeech results are reported as the CER results of net \textbar{} meeting subsets.
\item \label{note:audio-res:lyrics}
MIR-1K and Opencpop results are reported as CER.
\item \label{note:audio-res:audio-mc}
Audio MultiChallenge results are reported as the average pass rate.
\end{enumerate}
\end{minipage}
\end{table}

Although primarily centered on video understanding, the WeMM-Video family exhibits strong, well-rounded audio capabilities. Notably, WeMM-Video excels in general audio understanding. The 35B-A3B variant surpasses open-source baselines across widely used benchmarks like MMAU-Pro, MMAR, and MMSU, and performs on par with the proprietary Qwen3.5-Omni-Flash. Moreover, benefiting from targeted optimization, WeMM-Video leads on voice interaction and spoken instruction-following benchmarks, while also remaining highly competitive in music understanding as well as speech and lyrics recognition.


\subsection{Image Evaluation}

Table \ref{tab:img_res} presents the evaluation results on static image benchmarks across seven core dimensions: STEM and puzzle reasoning, general VQA, text and document understanding, spatial understanding, multi-image understanding, object counting, and multimodal instruction following. The evaluation suite encompasses 28 established benchmarks, including MMMU \cite{yue2024mmmu}, MMMU-Pro \cite{yue2025mmmu}, MathVision \cite{wang2024measuring}, MathVista \cite{lu2024mathvista}, DynaMath \cite{zou2025dynamath}, ZEROBench \cite{roberts2025zerobench}, ZEROBench\_sub \cite{roberts2025zerobench}, VisualPuzzles \cite{song2025visualpuzzles}, RealWorldQA \cite{realworldqa}, MMStar \cite{chen2024we}, MMBench$_{\text{EN-DEV-v1.1}}$ \cite{liu2024mmbench}, MMBench$_{\text{CN-DEV-v1.1}}$ \cite{liu2024mmbench}, SimpleVQA \cite{cheng2025simplevqa}, WorldVQA \cite{zhou2026worldvqa}, CharXiv RQ \cite{wang2024charxiv}, CC-OCR \cite{yang2025cc}, AI2D\_TEST \cite{kembhavi2016diagram}, MMLongBench-Doc \cite{ma2024mmlongbench}, OCRBench \cite{liu2024ocrbench}, DocVQA \cite{mathew2021docvqa}, ERQA \cite{team2025gemini}, CountBench \cite{paiss2023teaching}, RefCOCO (avg) \cite{yu2016modeling}, EmbSpatialBench \cite{du2024embspatial},
CV-Bench \cite{tong2024cambrian}, MIHBench \cite{li2025mihbench}, TallyQA \cite{acharya2019tallyqa}, and MMIFEval \cite{ding2025mm}.

\begin{table}[!htbp]
\centering
\caption{Evaluation results of the WeMM-Video family and baselines on image understanding benchmarks. Best results are in bold.}
\label{tab:img_res}
\resizebox{\columnwidth}{!}{%
\begin{tabular}{lccccccc}
\toprule
\multirow{2}{*}{\textbf{Benchmark}} & \textbf{Qwen3.5-Omni-} & \textbf{Qwen3.6-} &
\textbf{Qwen3.5-} & \multicolumn{4}{c}{\textbf{WeMM-Video-}} \\
& \textbf{Flash} & \textbf{35B-A3B} & \textbf{9B} & \textbf{2B (Thinking)} & \textbf{4B (Thinking)} & \textbf{9B (Thinking)} & \textbf{35B-A3B (Thinking)} \\
\midrule
\multicolumn{8}{c}{\textit{STEM/Puzzle}} \\
\midrule
MMMU & 76.9 & {\bfseries 81.7} & 78.4 & 61.2 & 78.4 & 78.9 & 81.1 \\
MMMU-Pro & 68.2 & {\bfseries 75.3} & 70.1 & 49.4 & 64.7 & 69.4 & 72.9 \\
MathVision & 65.4 & {\bfseries 80.4} & 78.9 & 49.2 & 73.3 & 76.0 & 79.6 \\
MathVista (mini) & 82.9 & 86.4 & 85.7 & 78.8 & 85.2 & 86.1 & {\bfseries 86.6} \\
DynaMath & 79.3 & 82.8 & {\bfseries 83.6} & 73.4 & 81.8 & 82.2 & 81.5 \\
ZEROBench & 1.0 & {\bfseries 5.6} & 3.0 & 0.4 & 0.8 & 2.0 & 3.8 \\
ZEROBench\_sub & 26.0 & {\bfseries 34.4} & 31.1 & 21.0 & 25.8 & 30.5 & 34.1 \\
VisualPuzzles & 44.9 & 58.1 & 55.1 & 39.6 & 53.4 & 56.8 & {\bfseries 60.1} \\

\midrule
\multicolumn{8}{c}{\textit{General VQA}} \\
\midrule
RealWorldQA & 77.5 & {\bfseries 85.3} & 80.3 & 73.6 & 80.1 & 80.3 & 83.4 \\
MMStar & 75.7 & {\bfseries 80.7} & 79.7 & 71.8 & 76.4 & 77.9 & 79.8 \\
$\text{MMBench}_\text{EN-DEV-v1.1}$ & 88.8 & {\bfseries 92.8} & 90.1 & 83.1 & 88.9 & 89.9 & 90.9 \\
$\text{MMBench}_\text{CN-DEV-v1.1}$ & 81.7 & 88.9 & 86.7 & 80.1 & 88.2 & 88.9 & {\bfseries 89.2} \\
SimpleVQA & 54.4 & 58.9 & 51.2 & 45.9 & 51.9 & 55.4 & {\bfseries 62.0} \\
WorldVQA & 13.6 & 19.9 & 15.7 & 14.1 & 20.3 & 23.6 & {\bfseries 28.6} \\

\midrule
\multicolumn{8}{c}{\textit{Text/Doc}} \\
\midrule
CharXiv RQ & 64.4 & {\bfseries 78.0} & 73.0 & 62.0 & 70.3 & 73.6 & 75.7 \\
CC-OCR & 80.8 & 81.9 & 79.3 & 79.8 & 82.9 & {\bfseries 84.7} & 84.2 \\
AI2D\_TEST & 89.0 & {\bfseries 92.7} & 90.2 & 83.2 & 88.4 & 89.6 & 90.6 \\
MMLongBench-Doc & 53.6 & {\bfseries 61.2} & 57.7 & 42.8 & 55.5 & 56.1 & 60.2 \\
OCRBench & 89.1 & 87.9 & 89.2 & 81.8 & 85.7 & 89.6 & {\bfseries 90.2} \\
DocVQA & 95.3 & 97.3 & 97.2 & 95.1 & 96.5 & {\bfseries 97.5} & {\bfseries 97.5} \\

\midrule
\multicolumn{8}{c}{\textit{Spatial}} \\
\midrule
ERQA & 50.0 & 59.8 & 55.5 & 43.8 & 53.0 & 54.5 & {\bfseries 61.8} \\
RefCOCO (avg) & {\bfseries 92.6} & 92.0 & 89.7 & 63.7 & 86.3 & 89.0 & 89.5 \\
EmbSpatialBench & 82.7 & {\bfseries 84.3} & 83.0 & 79.0 & 83.3 & 82.8 & 83.9 \\
CV-Bench & 86.6 & 87.8 & 87.2 & 84.2 & 88.6 & 88.6 & {\bfseries 89.2} \\

\midrule
\multicolumn{8}{c}{\textit{Multi-Image}} \\
\midrule
MIHBench & 61.8 & 81.3 & 81.8 & 80.8 & 81.5 & 82.9 & {\bfseries 83.1} \\

\midrule
\multicolumn{8}{c}{\textit{Counting}} \\
\midrule
CountBench & 88.2 & 96.3 & {\bfseries 97.2} & 83.9 & 93.3 & 94.3 & 96.3 \\
TallyQA & 81.2 & 79.8 & 79.8 & 78.8 & 80.4 & {\bfseries 81.5} & 80.7 \\

\midrule
\multicolumn{8}{c}{\textit{Instruction-Following}} \\
\midrule
MMIFEval & 67.0 & 81.1 & 79.5 & 58.8 & 74.6 & 80.8 & {\bfseries 81.4} \\

\bottomrule
\end{tabular}%
}
\end{table}

Overall, the WeMM-Video family effectively preserves the strong image perception and reasoning capabilities of its vision-language backbones from omni-modal training. Specifically, WeMM-Video-35B-A3B and WeMM-Video-9B comprehensively outperform Qwen3.5-Omni-Flash across nearly all evaluated benchmarks, while performing on par with their respective Qwen3.6-35B-A3B and Qwen3.5-9B backbones.

\subsection{Text Evaluation}
Table \ref{tab:text_res} presents the evaluation results on pure text benchmarks across five core dimensions: general knowledge, instruction following, long-context understanding, STEM problem solving, and mathematical reasoning. The evaluation suite encompasses 12 established benchmarks, including MMLU-Pro \cite{wang2024mmlu}, MMLU-Redux \cite{gema2025we}, SuperGPQA \cite{du2026supergpqa}, C-Eval \cite{huang2023c}, IFEval \cite{zhou2023instruction}, IFBench \cite{pyatkin2026generalizing}, CFBench \cite{zhang2025cfbench}, AA-LCR \cite{artificialanalysis2025lcr}, LongBench v2 \cite{bai2025longbench}, GPQA \cite{rein2023gpqa}, HMMT Nov 25 \cite{dekoninck2026matharena}, and IMOAnswerBench \cite{luong2025towards}.

\begin{table}[!htbp]
\centering
\caption{Evaluation results of the WeMM-Video family and baselines on pure text benchmarks. Best results are in bold.}
\label{tab:text_res}
\resizebox{\columnwidth}{!}{%
\begin{tabular}{lccccccc}
\toprule
\multirow{2}{*}{\textbf{Benchmark}} & \textbf{Qwen3.5-Omni-} & \textbf{Qwen3.6-} & \textbf{Qwen3.5-} & \multicolumn{4}{c}{\textbf{WeMM-Video-}} \\
 & \textbf{Flash} & \textbf{35B-A3B} & \textbf{9B} & \textbf{2B (Thinking)} & \textbf{4B (Thinking)} & \textbf{9B (Thinking)} & \textbf{35B-A3B (Thinking)} \\
\midrule
\multicolumn{8}{c}{\textit{Knowledge}} \\
\midrule
MMLU-Pro & 79.9 & {\bfseries 85.2} & 82.5 & 63.7 & 78.3 & 81.6 & 84.7 \\
MMLU-Redux & 90.0 & 93.3 & 91.1 & 75.7 & 88.7 & 90.5 & {\bfseries 93.7} \\
SuperGPQA & 54.9 & {\bfseries 64.7} & 58.2 & 36.3 & 52.4 & 56.7 & 63.4 \\
C-Eval & 86.0 & 90.0 & 88.2 & 72.0 & 84.8 & 87.6 & {\bfseries 90.4} \\

\midrule
\multicolumn{8}{c}{\textit{Instruction-Following}} \\
\midrule
IFEval & 85.2 & 90.9 & 91.5 & 84.3 & 91.3 & {\bfseries 93.2} & {\bfseries 93.2} \\
IFBench & 38.4 & 63.7 & 64.5 & 45.3 & 63.7 & {\bfseries 70.7} & {\bfseries 70.7} \\
CFBench & 83.4 & 87.3 & 85.3 & 77.6 & 86.0 & 86.9 & {\bfseries 87.7} \\

\midrule
\multicolumn{8}{c}{\textit{Long Context}} \\
\midrule
AA-LCR & 46.0 & 60.0 & 63.0 & 14.0 & 56.0 & 59.0 & {\bfseries 64.0} \\
LongBench v2 & 46.4 & {\bfseries 59.6} & 55.2 & 33.8 & 53.5 & 54.7 & 58.1 \\

\midrule
\multicolumn{8}{c}{\textit{STEM}} \\
\midrule
GPQA & 76.4 & {\bfseries 86.0} & 81.7 & 56.1 & 73.2 & 77.8 & 83.8 \\


\midrule
\multicolumn{8}{c}{\textit{Reasoning}} \\
\midrule
HMMT Nov 25 & 59.0 & 89.1 & 82.9 & 26.7 & 75.8 & 75.0 & {\bfseries 90.0} \\
IMOAnswerBench & 51.5 & {\bfseries 78.9} & 67.8 & 22.5 & 58.8 & 62.8 & 74.8 \\


\bottomrule
\end{tabular}%
}
\end{table}

Consistent with the image evaluation findings, incorporating audio-visual modalities into WeMM-Video preserves the core linguistic capabilities of the base models. Across all evaluated dimensions, WeMM-Video-35B-A3B and WeMM-Video-9B substantially outperform Qwen3.5-Omni-Flash, while performing on par with their respective Qwen3.6-35B-A3B and Qwen3.5-9B backbones and achieving further improvements in instruction following.

\section{Conclusion}

In this report, we present WeMM-Video, a family of Chinese-English bilingual omni-modal foundation models spanning 2B, 4B, 9B, and 35B-A3B parameter scales. Centered on omni-modal video understanding and reasoning, WeMM-Video unifies text, image, audio, and video processing within a hybrid-thinking paradigm through progressive three-stage pre-training and a multi-stage post-training pipeline comprising multi-teacher OPD, anti-repetition MrDPO, and specialist RL. Comprehensive evaluations demonstrate that our flagship WeMM-Video-35B-A3B establishes a new open-source state of the art in omni-modal video understanding and rivals leading proprietary models, while our lightweight variants exhibit strong cross-scale efficiency. Beyond joint audio-visual comprehension, the WeMM-Video family also delivers leading performance across audio, silent video, image, and text modalities. We publicly release the model weights along with the complete evaluation and reproduction pipeline, and we hope WeMM-Video will serve as a solid foundation for future research and real-world omni-modal applications.

\clearpage
\bibliography{main}
\bibliographystyle{plainnat}
\end{document}

%% file: math_commands.tex
\usepackage{amsmath,amsfonts,bm}

\def\eqref#1{equation~\ref{#1}}

\def\1{\bm{1}}

\DeclareMathAlphabet{\mathsfit}{\encodingdefault}{\sfdefault}{m}{sl}
\SetMathAlphabet{\mathsfit}{bold}{\encodingdefault}{\sfdefault}{bx}{n}

